\documentclass[runningheads]{llncs}
\usepackage{eccv}
\usepackage{eccvabbrv}
\usepackage{graphicx}
\graphicspath{{figure/}}
\usepackage{booktabs}
\usepackage{multirow}
\usepackage[accsupp]{axessibility}
\usepackage[pagebackref,breaklinks,colorlinks,citecolor=eccvblue]{hyperref}
\usepackage{hyperref}
\usepackage{orcidlink}

\begin{document}
\title{Oscillating Dispersion for Maximal Light-throughput Spectral Imaging} 

\author{Jiuyun Zhang\inst{1} \and
Zhan Shi\inst{1} \and
Linsen Chen\inst{1}\thanks{Corresponding author.} \and Xun Cao\inst{1,2}}

\authorrunning{J.~Zhang et al.}

\institute{Nanjing University, Nanjing, China \and Key Laboratory of Optoelectronic Devices and Systems with Extreme Performances of MOE, Nanjing University, Nanjing, China \\
\email{\{jiuyun\_zhang, shizhan\}@smail.nju.edu.cn,\{chenls, caoxun\}@nju.edu.cn}}

\maketitle

\begin{abstract}
  Existing computational spectral imaging systems typically rely on coded aperture and beam splitters that block a substantial fraction of incident light, degrading reconstruction quality under light-starved conditions. To address this limitation, we develop the Oscillating Dispersion Imaging Spectrometer (ODIS), which for the first time achieves near-full light throughput by axially translating a disperser between the conjugate image plane and a defocused position, sequentially capturing a panchromatic (PAN) image and a dispersed measurement along a single optical path. We further propose a PAN-guided Dispersion-Aware Deep Unfolding Network (PDAUN) that recovers high-fidelity spectral information from maskless dispersion under PAN structural guidance. Its data-fidelity step derives an FFT-Woodbury preconditioned solver by exploiting the cyclic-convolution property of the ODIS forward model, while a Dispersion-Aware Deformable Convolution module (DADC) corrects sub-pixel spectral misalignment using PAN features. Experiments show state-of-the-art performance on standard benchmarks, and cross-system comparisons confirm that ODIS yields decisive gains under low illumination. High-fidelity reconstruction is validated on a physical prototype.
\end{abstract}

\section{Introduction}
\label{sec:intro}

\begin{figure}[tb]
  \centering
  \includegraphics[width=\linewidth]{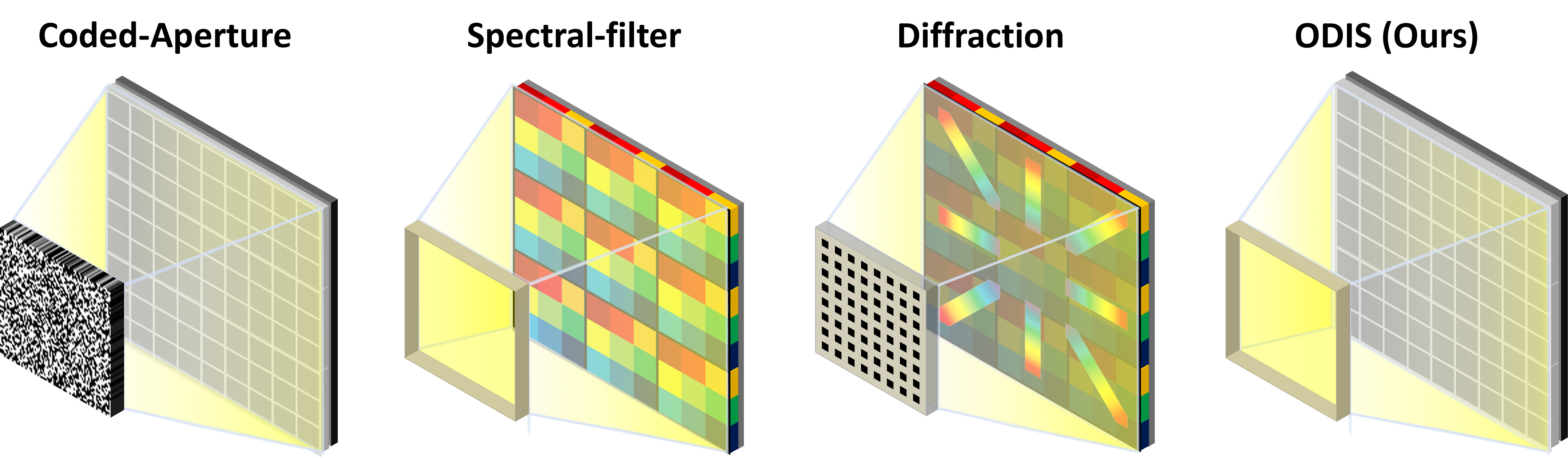}
  \caption{Depiction of different computational spectral imaging systems. From left to right: coded-aperture systems modulate light with an amplitude mask; spectral-filter-based and diffraction-based systems encode spectral information via wavelength-selective filters or diffractive elements, respectively; ODIS encodes spectra through axial prism translation with a grayscale camera.}
  \label{fig:teaser}
  
  \captionof{table}{Comparison of representative computational spectral imaging architectures.}
  \label{tab:system-compare}
  \centering
  \footnotesize
  \setlength{\tabcolsep}{2.5pt}
  \begin{tabular}{@{}lccccccc@{}}
    \toprule
    Property & CASSI & HVIS & DCCHI & SFA & ADIS & DOE & ODIS \\
    \midrule
    Type & Coded & Coded & Coded & Filter & Diffr. & Diffr. & -- \\
    Cameras & gray & gray+RGB & gray$\times$2 & gray & RGB & RGB & gray \\
    Throughput & ${\sim}50\%$ & ${\sim}1/2C$ & ${\sim}25\%$ & ${\sim}1/C$ & ${\sim}25\%$ & ${\sim}33\%$ & ${\sim}100\%$ \\
    \bottomrule
  \end{tabular}
  \vspace{-1.0em}
\end{figure}

Spectral information encodes rich physicochemical properties of scenes and underpins a broad spectrum of applications, from remote sensing and medical diagnostics to industrial inspection~\cite{bioucas2013hyperspectral,calin2014hyperspectral,lin2002reuven,shimoni2019hyperspectral}. Computational spectral imaging (CSI) overcomes the efficiency and compactness limitations of traditional wavelength-scanning spectrometers through joint optical encoding and computational reconstruction design, and has become the prevailing paradigm for acquiring hyperspectral images~\cite{cao2016computational,lam2015computational,yuan2021snapshot}. However, existing CSI systems universally rely on optical coding elements, such as coded-aperture~\cite{arce2013compressive,gehm2007single,wagadarikar2008single}, spectral filter arrays~\cite{lapray2014multispectral,mihoubi2017multispectral,wang2007concept,tittl2018imaging}, and diffractive optical elements~\cite{jeon2019compact}, that inevitably block a substantial fraction of the incident light, sacrificing system throughput. This throughput loss directly degrades the signal-to-noise ratio (SNR) of measurements, and reconstruction quality deteriorates sharply under photon-starved conditions, severely limiting the practical deployment of CSI in low-light scenarios such as fluorescence microscopy~\cite{haraguchi2002spectral,fereidouni2012spectral}, nighttime environmental monitoring~\cite{miller2009dynamic}, and surveillance under poor illumination~\cite{kim2019deep}.

The coded aperture snapshot spectral imager (CASSI)~\cite{wagadarikar2008single,gehm2007single}, including single-disperser (SD)~\cite{wagadarikar2008single} and dual-disperser (DD)~\cite{gehm2007single} types, places a random binary mask in the optical path that blocks roughly 50\% of the incident light. To obtain non-aliased spectral sampling, PMVIS~\cite{cao2011prism} employs a sparse mask that transmits only one spectral channel per pixel, retaining merely $1/C$ of the incident energy; conversely, its variant DDN~\cite{chen2023notch} adopts a notch mask that blocks only $1/C$ of the light to preserve the maximal mask throughput, yet its reliance on an RGB sensor further discards approximately two-thirds of the transmitted photons through Bayer filtering. To acquire a complementary panchromatic (PAN) reference, several designs~\cite{wang2015dual,wang2015high} augment coded systems with a beam splitter to form a dual-camera architecture; yet the splitter itself further halves per-arm throughput, and the two sensors introduce the additional burden of precise spatial co-registration. Spectral filter array (SFA) approaches integrate narrowband filters at each pixel for compact snapshot acquisition, yet custom fabrication is costly and per-pixel filtering limits effective throughput. As a promising alternative, coding-free approaches attempt to bypass amplitude masks entirely. Diffraction-based methods~\cite{lv2023aperture} exploit diffractive optical elements (DOEs) or gratings for spectral encoding with no inherent amplitude loss, and dispersion-based methods~\cite{zhao2019spectral} pioneer the extraction of spectral information from dispersive blur with high light efficiency. Unfortunately, current diffraction systems still couple diffractive elements with RGB mosaic sensors, limiting effective throughput to roughly 25--33\%, while dispersion-based designs rely on a dual-path layout together with an auxiliary edge mask for reconstruction~\cite{zhao2019spectral}. Despite these diverse encoding strategies, a fundamental barrier persists: no existing system simultaneously achieves full light throughput, single-camera simplicity, and inherent spatial registration.

In this paper, we develop the Oscillating Dispersion Imaging Spectrometer (ODIS), which for the first time achieves near-full light throughput in computational spectral imaging. The key insight is that translating a dispersive element within a single optical path, instead of splitting the beam into two arms, allows one sensor to sequentially acquire a dispersed measurement and a PAN reference, preserving full throughput while eliminating spatial co-registration. Specifically, ODIS axially translates a dispersive prism between the conjugate image plane and a defocused position, capturing a PAN image in the former and a dispersed-blur measurement in the latter, as illustrated in \cref{fig:system}. This mechanism requires only millimeter-scale travel along a single optical path with a single camera, thereby achieving: (i)~near-100\% light throughput without mask or beam splitter; (ii)~single-sensor acquisition for a compact, low-cost system; and (iii)~inherent spatial registration between the dispersed measurement and the PAN image without any calibration. As a trade-off, ODIS requires two sequential exposures, partially sacrificing temporal resolution for dynamic scene acquisition. \cref{tab:system-compare} contrasts ODIS with representative CSI architectures. Controlled cross-system simulations corroborate that the full-throughput advantage of ODIS translates into decisive reconstruction gains as illumination decreases (\cref{sec:cross-system}).

Furthermore, existing reconstruction algorithms are designed around mask-modulated forward models and cannot fully exploit the mask-free dispersion geometry of ODIS. To address this, we propose a PAN-guided Dispersion-Aware Deep Unfolding Network (PDAUN) tailored to the joint dispersion-plus-PAN forward model of ODIS. Within the ADMM unfolding framework, the data-fidelity step derives an FFT-Woodbury preconditioned conjugate-gradient (PCG) solver that exploits the cyclic-convolution property of the ODIS dispersion operator for efficient, physics-faithful updates; in the denoising prior step, a Dispersion-Aware Deformable Convolution module (DADC) leverages PAN features to correct sub-pixel spectral misalignment. Extensive experiments demonstrate that PDAUN achieves state-of-the-art performance on standard benchmarks, and a physical prototype further validates high-fidelity spectral recovery in real-world settings.

In summary, this paper makes the following contributions. We present a system--algorithm co-design that, for the first time, enables near-full-throughput computational spectral imaging together with a dedicated reconstruction framework:
\begin{itemize}
  \item We develop ODIS, a compact single-camera spectral imaging system that achieves near-full light throughput without any coded mask or beam splitter, and provides inherent spatial registration between measurements.
  \item We propose PDAUN, an ADMM deep unfolding network tailored to the joint dispersion--PAN forward model of ODIS for high-fidelity spectral reconstruction. An FFT-Woodbury preconditioned CG solver exploits the cyclic-convolution property of the dispersion operator for efficient data-fidelity updates, while DADC leverages PAN guidance for sub-pixel dispersion correction.
  \item Comprehensive experimental validation including controlled cross-system comparisons that demonstrate the decisive advantage of full throughput under low illumination, state-of-the-art results on standard benchmarks, ablation studies, and physical prototype verification.
\end{itemize}

\section{Related Work}
\label{sec:related}

\noindent\textbf{Coded-aperture systems.}
Amplitude-coded systems that spatially modulate the incoming light with binary masks or filter arrays dominate current computational spectral imaging. SD-CASSI~\cite{wagadarikar2008single} places a single disperser after a coded aperture for spatially encoded compressive sensing; DD-CASSI~\cite{gehm2007single} sandwiches the mask between two dispersers to add spectral-domain encoding; DDN~\cite{chen2023notch} replaces the random binary mask with a sparse notch mask paired with an RGB camera, converting the system from multiplexed measurement to direct spectral sampling; and PMVIS~\cite{cao2011prism} combines a sparse mask array with a prism so that each pixel captures only one wavelength. To supplement the spatial information lost through coded modulation, dual-camera variants have been developed: DCCHI~\cite{wang2015dual} augments CASSI with a beam-splitter-coupled panchromatic camera, and HVIS~\cite{cao2011high} pairs PMVIS with an RGB camera to provide a high-resolution spatial reference. However, the beam splitter in these dual-path designs further halves per-arm throughput and introduces cross-camera spatial co-registration burdens. The common bottleneck across all coded-aperture architectures remains the amplitude mask, which inevitably blocks a substantial fraction of incident light (\cref{tab:system-compare}). ODIS removes the coding mask entirely, using the dispersive prism as the sole encoding element to achieve near-100\% light throughput with single-camera compactness and inherent spatial registration.

\noindent\textbf{Spectral-filter-based systems.}
A separate class of methods integrates spectral-selective filtering devices on the sensor plane. Broadband spectral filter array (SFA) approaches place a narrow-band thin-film filter at each pixel and recover spatial resolution through computational reconstruction; although throughput can exceed 50\%, the custom fabrication of filter arrays is costly and difficult to scale~\cite{lapray2014multispectral}. RGB-camera-based spectral reconstruction methods exploit the broadband three-channel response of Bayer filter arrays and estimate multi-channel spectral information via learned models, yet the Bayer array itself blocks approximately two-thirds of the incident light. ODIS uses a monochrome sensor with no spectral filtering devices, so every pixel receives the full spectral content of the scene without filter-induced throughput loss.

\noindent\textbf{Diffraction-based systems.}
Diffraction-based methods exploit wavefront phase modulation of diffractive optical elements (DOEs) or gratings for spectral separation or encoding. The Computed Tomography Imaging Spectrometer (CTIS)~\cite{descour1995computed} interprets spectral information as multi-angle projections via a cross grating, but sacrifices substantial spatial resolution due to cone loss. ADIS~\cite{lv2023aperture} achieves compact snapshot spectral imaging through diffractive multiplexing with a mosaic sensor; however, its mosaic encoding remains inherently pixel-level spectral sampling, entailing a fundamental spatial--spectral resolution trade-off. DOE-based designs with optimized phase masks enable joint control of dispersion and focus~\cite{katkovnik2018optimization}, yet their long design cycles and high fabrication costs limit reproducibility. ODIS employs a standard Amici refractive prism rather than diffractive elements---an off-the-shelf component that is easy to procure and low in cost.

\noindent\textbf{Reconstruction algorithms.}
Spectral reconstruction has progressed from early hand-crafted-prior iterative methods~\cite{zhao2019spectral,yuan2016generalized,liu2018rank} to end-to-end deep networks~\cite{meng2020end,huang2021deep,hu2022hdnet,cai2022mask,cai2022coarse,yao2024specat}, and further to deep unfolding networks that embed a physical forward model into learnable stages, with DAUHST~\cite{cai2022degradation} and In2SET~\cite{wang2024in2set} representing the state of the art for single- and dual-camera CASSI, respectively. However, the data-fidelity steps of all these methods rely on the spatial modulation provided by a coded mask; removing the mask causes them to degenerate into scalar scaling, losing spatial discrimination. Pansharpening~\cite{loncan2015hyperspectral} and hybrid-camera fusion~\cite{gu2025hgsfusion} leverage a high-resolution auxiliary image for spectral reconstruction, yet they assume a separate auxiliary sensor and require cross-modal registration. PDAUN is the first deep unfolding network designed for the joint mask-free dispersion-plus-same-sensor PAN forward model of ODIS.

\section{ODIS: System Design}
\label{sec:system}

\begin{figure}[tb]
  \centering
  \includegraphics[width=\linewidth]{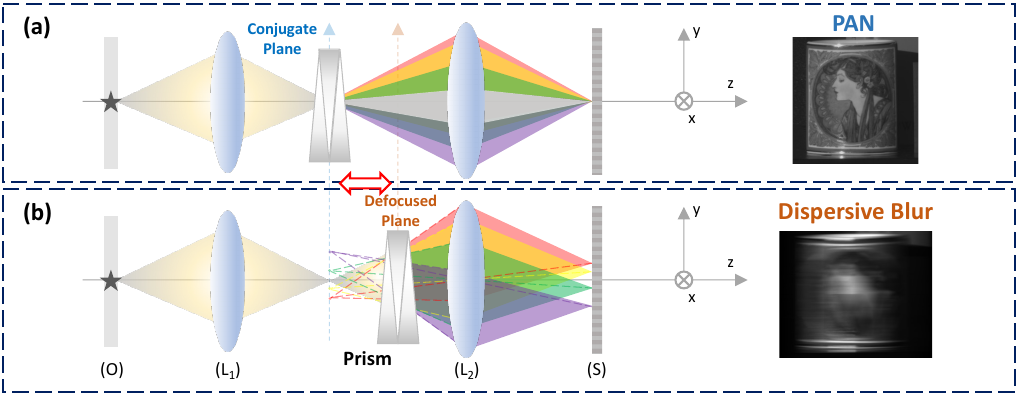}
  \caption{Optical layout of ODIS. (a) With the Amici prism at the image-conjugate plane, all wavelengths from a scene point converge to a common sensor location, producing a sharp PAN image. (b) Translating the prism along the optical axis to the defocused plane causes each wavelength to focus at a distinct lateral position, yielding a spectrally dispersed measurement.}
  \label{fig:system}
\end{figure}

The central idea behind ODIS is to oscillate between two acquisition modes by axially translating a dispersive prism within the single optical path (\cref{fig:system}).

\noindent\textbf{Image-conjugate Plane.}
When the prism sits at the image-conjugate plane of the relay optics, all wavelengths from a given scene point converge onto the same sensor location after refraction. The sensor therefore records a PAN image $\mathbf{Y}_{\text{pan}} \in \mathbb{R}^{H \times W}$ that preserves full spatial resolution.

\noindent\textbf{Defocused Plane.}
Translating the prism by a few millimetres along the optical axis moves it to a defocused plane. Different wavelengths now arrive at distinct sensor columns, producing a controlled spectral dispersion. The sensor records a dispersed-blur measurement $\mathbf{Y}_{\text{disp}} \in \mathbb{R}^{H \times (W + d(C-1))}$, where $d$ denotes the dispersion step in pixels per spectral channel and $C$ is the number of channels. With the current K9--ZF88 Amici prism, a travel of only 2\,mm yields a dispersion step of 1\,pixel per 10\,nm, well within the range of compact linear actuators.

Because both measurements traverse the same single optical path and share the same sensor, ODIS achieves three properties simultaneously: (i)~near-100\% light throughput, as no coded mask or beam splitter is required; (ii)~single-camera compactness; and (iii)~inherent spatial registration between the dispersed and PAN measurements. As a trade-off, the two sequential exposures introduce a finite temporal gap between the dispersed and PAN acquisitions, partially sacrificing temporal resolution when imaging dynamic scenes. The supplementary material shows that the deviation introduced by the finite prism thickness amounts to a small axial shift of the focal plane, which does not affect the validity of the forward model presented above. \cref{tab:system-compare} in \cref{sec:intro} contrasts ODIS with representative CSI architectures.

\noindent\textbf{Forward model.}
Let $\mathbf{X} \in \mathbb{R}^{H \times W \times C}$ denote the Hyperspectral Image (HSI), where $H$, $W$, and $C$ are the spatial height, width, and number of spectral channels, respectively. At the defocused position, the dispersive prism shifts the $c$-th spectral channel by $d(c{-}1)$ pixels along the $y$-axis ($d$: dispersion step per channel). The dispersed measurement recorded at detector pixel $(x,y)$ is
\begin{equation}
  \label{eq:disp}
  Y_{\text{disp}}(x, y) = \sum_{c=1}^{C} X\!\bigl(x,\; y{-}d(c{-}1),\; c\bigr) + N_1(x,y).
\end{equation}
Every detector pixel receives the full sum of all spectral channels at their respective dispersion-shifted positions, where $N_1$ and $N_2$ denote the measurement noise of the dispersed and PAN acquisitions, respectively. At the image-conjugate position, no spectral dispersion occurs and the sensor records a PAN image that integrates all channels at each spatial location:
\begin{equation}
  \label{eq:pan}
  Y_{\text{pan}}(x, y) = \sum_{c=1}^{C} X(x,\; y,\; c) + N_2(x,y).
\end{equation}

For compact notation, vectorising the data cube as $\mathbf{x} = \mathrm{vec}(\mathbf{X}) \in \mathbb{R}^{HWC}$ and stacking both measurements yields a unified matrix-vector form:
\begin{equation}
  \label{eq:joint}
  \mathbf{y} = \mathbf{A}\,\mathbf{x} + \mathbf{n}, \qquad
  \mathbf{A} = \begin{bmatrix} \mathbf{A}_{\text{disp}} \\ \mathbf{A}_{\text{pan}} \end{bmatrix},
\end{equation}
where $\mathbf{A}_{\text{disp}}$ encodes the channel-wise uniform shift-and-sum of \cref{eq:disp} and $\mathbf{A}_{\text{pan}}$ encodes the spectral integration of \cref{eq:pan}.

\noindent\textbf{Cyclic-convolution structure.}
As shown in \cref{eq:disp}, the dispersion operator applies an identical set of uniformly spaced shifts to every spatial row, an operation equivalent to a cyclic convolution along the spectral--spatial dispersion axis. Consequently, $\mathbf{A}_{\text{disp}}^{T}\mathbf{A}_{\text{disp}}$ admits a block-circulant structure that can be diagonalised efficiently in the Fourier domain. This property is the mathematical foundation of the FFT-Woodbury preconditioned solver developed in \cref{sec:fftpcg}.

\begin{figure}[tb]
  \centering
  \includegraphics[width=\linewidth]{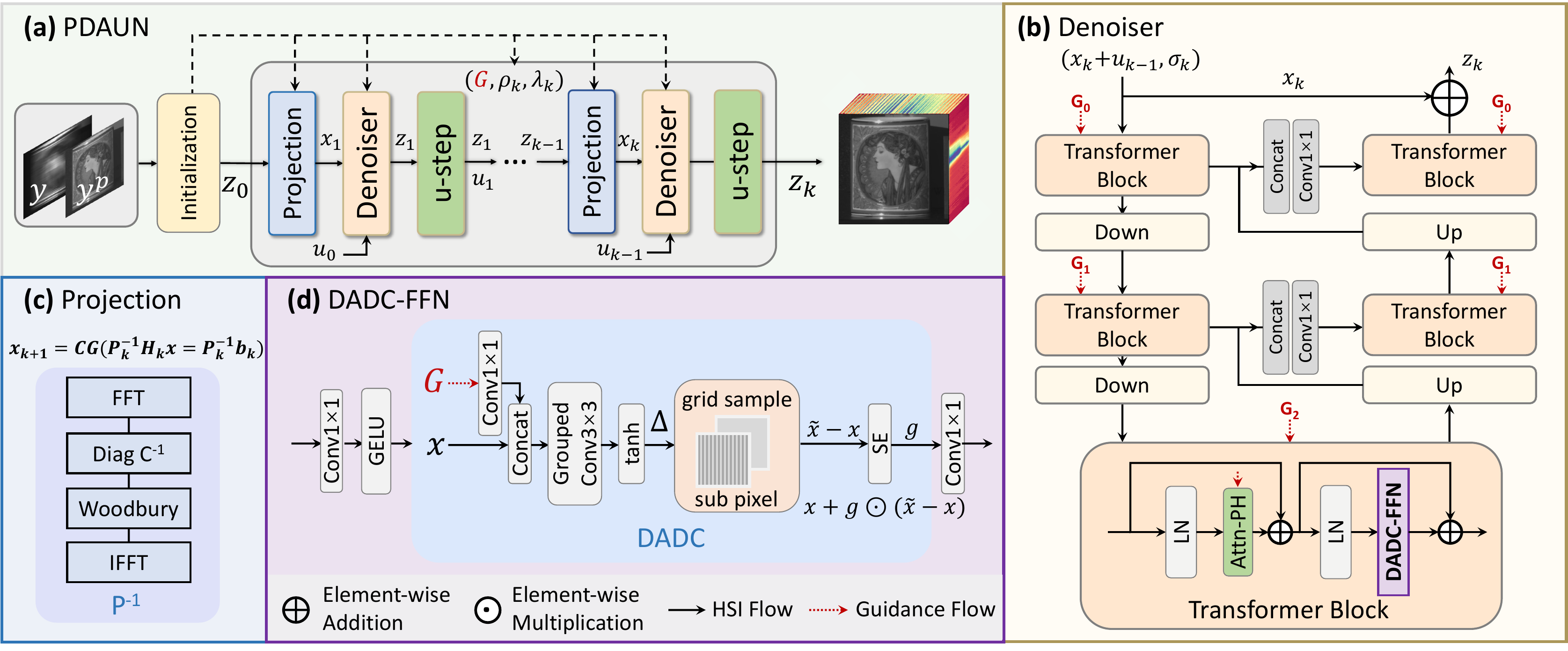}
  \caption{Overview of PDAUN. (a) PDAUN architecture with $K$ unfolding stages. (b) Denoiser with three-layer U-shaped structure. (c) FFT-Woodbury preconditioned conjugate-gradient solver. (d) Dispersion-Aware Deformable Convolution (DADC) module.}
  \label{fig:architecture}
\end{figure}

\section{PDAUN: Algorithm Design}
\label{sec:algorithm}

Although the ODIS forward model is linear, directly applying existing CSI reconstruction pipelines is suboptimal. Traditional iterative methods~\cite{zhao2019spectral,yuan2016generalized,liu2018rank} rely on hand-crafted priors and do not scale to the joint dispersion-plus-PAN observation. Pure data-driven networks~\cite{meng2020end,huang2021deep,hu2022hdnet,cai2022mask,cai2022coarse,yao2024specat} lack explicit enforcement of the ODIS physical constraints. Existing deep unfolding methods~\cite{cai2022degradation,wang2024in2set} are built around mask-modulated forward models and perform suboptimally under the mask-free ODIS geometry. These gaps motivate a dedicated unfolding algorithm tailored to ODIS.

\subsection{ADMM Deep Unfolding Framework}

Let $\mathbf{x}$ denote the vectorized HSI cube, and let $\mathbf{A}_{\text{disp}}$, $\mathbf{A}_{\text{pan}}$ be the two forward operators in \cref{sec:system}. PDAUN solves
\begin{equation}
  \min_{\mathbf{x}}\; \frac{1}{2}\lVert \mathbf{A}_{\text{disp}}\mathbf{x}-\mathbf{y}_{\text{disp}} \rVert_2^2
  + \frac{\lambda_k}{2}\lVert \mathbf{A}_{\text{pan}}\mathbf{x}-\mathbf{y}_{\text{pan}} \rVert_2^2
  + \mathcal{R}(\mathbf{x}),
\end{equation}
where $\mathcal{R}(\mathbf{x})$ denotes a regularization prior on the HSI cube. By introducing an auxiliary variable $\mathbf{z}$ and scaled dual variable $\mathbf{u}$, the $k$-th ADMM stage decomposes into three steps, as is shown in \cref{fig:architecture}. The data-fidelity step ($\mathbf{x}$-subproblem) enforces consistency with both the dispersed and PAN measurements:
\begin{equation}
  \mathbf{x}_{k+1}=\arg\min_{\mathbf{x}}\; \frac{1}{2}\lVert \mathbf{A}_{\text{disp}}\mathbf{x}-\mathbf{y}_{\text{disp}} \rVert_2^2
  + \frac{\lambda_k}{2}\lVert \mathbf{A}_{\text{pan}}\mathbf{x}-\mathbf{y}_{\text{pan}} \rVert_2^2
  + \frac{\rho_k}{2}\lVert \mathbf{x}-\mathbf{z}_{k}+\mathbf{u}_{k} \rVert_2^2,
\end{equation}
where $\lambda_k$ balances the relative weight of the PAN term and $\rho_k$ couples this step to the prior. The prior step ($\mathbf{z}$-subproblem) amounts to a PAN-guided denoising problem with equivalent noise level $\sigma_k$:
\begin{equation}
  \mathbf{z}_{k+1}=\mathcal{D}_{\theta_k}\!\left(\mathbf{x}_{k+1}+\mathbf{u}_{k},\,\mathbf{y}_{\text{pan}},\,\sigma_k\right),
  \quad \sigma_k=1/\sqrt{\rho_k},
\end{equation}
where the PAN image $\mathbf{y}_{\text{pan}}$ provides high-resolution structural guidance to the learned denoiser $\mathcal{D}_{\theta_k}$ (detailed in \cref{sec:dadc}). The dual-variable update accumulates the constraint violation between $\mathbf{x}$ and $\mathbf{z}$, steering subsequent iterations toward consensus:
\begin{equation}
  \mathbf{u}_{k+1}=\mathbf{u}_{k}+\mathbf{x}_{k+1}-\mathbf{z}_{k+1}.
\end{equation}
PDAUN unrolls these updates into $K$ stages, where each stage has an independent prior network $\mathcal{D}_{\theta_k}$.

\subsection{FFT-Woodbury Preconditioned CG for the x-step}
\label{sec:fftpcg}

The $\mathbf{x}$-subproblem leads to the linear system
\begin{equation}
  \label{eq:xstep}
  \bigl(\mathbf{A}_{\text{disp}}^\top\mathbf{A}_{\text{disp}}
  +\lambda_k\mathbf{A}_{\text{pan}}^\top\mathbf{A}_{\text{pan}}
  +\rho_k\mathbf{I}\bigr)\mathbf{x}
  =\mathbf{A}_{\text{disp}}^\top\mathbf{y}_{\text{disp}}
  +\lambda_k\mathbf{A}_{\text{pan}}^\top\mathbf{y}_{\text{pan}}
  +\rho_k(\mathbf{z}_{k}-\mathbf{u}_{k}).
\end{equation}
Hereafter we write \cref{eq:xstep} compactly as $\mathbf{H}_k\mathbf{x}=\mathbf{b}_k$.
For ODIS, channel-wise dispersion is uniform-shift-and-sum, so $\mathbf{A}_{\text{disp}}^\top\mathbf{A}_{\text{disp}}$ is well approximated by a block-circulant operator under cyclic boundary modeling, diagonalizable in the Fourier domain. Meanwhile, $\mathbf{A}_{\text{pan}}^\top\mathbf{A}_{\text{pan}}$ is rank-1 in the spectral dimension. Defining the block-circulant component $\mathbf{C}\triangleq\mathbf{A}_{\text{disp}}^{\top}\mathbf{A}_{\text{disp}}+\rho_k\mathbf{I}$, whose inverse is applied element-wise in the Fourier domain, the Woodbury identity yields
\begin{equation}
  \label{eq:precond}
  \mathbf{P}_k^{-1} = \mathbf{C}^{-1}
    - \lambda_k\,\mathbf{C}^{-1}\mathbf{A}_{\text{pan}}^{\top}
      \!\bigl(\mathbf{I}+\lambda_k\,\mathbf{A}_{\text{pan}}\mathbf{C}^{-1}\mathbf{A}_{\text{pan}}^{\top}\bigr)^{-1}
      \!\mathbf{A}_{\text{pan}}\mathbf{C}^{-1},
\end{equation}
where the rank-1 spectral structure of $\mathbf{A}_{\text{pan}}$ reduces the inner inverse to a per-pixel scalar division. Each application of $\mathbf{P}_k^{-1}$ requires only two FFT passes, giving $\mathcal{O}(HWC\log(WC))$ complexity.

Importantly, the practical imaging boundary is not strictly cyclic (linear shift with finite support), so $\mathbf{P}_k^{-1}$ is not an exact inverse of the true normal matrix $\mathbf{H}_k$. The data-fidelity solution is therefore obtained via preconditioned conjugate gradients:
\begin{equation}
  \label{eq:pcg}
  \mathbf{x}_{k+1} = \operatorname{CG\text{-}Solve}\,\bigl(\mathbf{P}_k^{-1}\mathbf{H}_k\,\mathbf{x}=\mathbf{P}_k^{-1}\mathbf{b}_k;\;T\text{ iterations}\bigr),
\end{equation}
where $\mathbf{P}_k^{-1}\mathbf{H}_k\approx\mathbf{I}$, so the condition number is close to~1 and very few iterations suffice for high-accuracy convergence. To reduce memory overhead during training, the PCG module uses implicit differentiation in the backward pass. The detailed derivation is provided in the supplementary material.

\subsection{Dispersion-Aware Deformable Convolution}
\label{sec:dadc}

The ODIS dispersion operator shifts each spectral channel by a different number of pixels, causing channel-dependent sub-pixel misalignment that varies with spatial content. Standard depth-wise convolutions apply spatially fixed kernels and cannot adapt to these shifts. DADC addresses this by using the PAN image, which is captured on the same sensor without dispersion, as a spatial anchor. Concretely, the multi-scale PAN feature $\mathbf{G}$ is projected and concatenated with the intermediate HSI feature $\mathbf{X}$ to predict per-channel 2-D spatial offsets:
\begin{equation}
  \label{eq:offset}
  \boldsymbol{\Delta} = s \cdot \tanh\!\bigl(\mathrm{GroupConv}([\mathrm{Proj}(\mathbf{G}),\;\mathbf{X}])\bigr),
\end{equation}
where the learnable scalar $s$ bounds the maximum shift magnitude. Each channel is then resampled at its predicted sub-pixel location via differentiable bilinear interpolation, yielding $\tilde{\mathbf{X}}$. A channel-wise gate fuses the shifted and original features:
\begin{equation}
  \label{eq:gate}
  \mathbf{g} = \sigma\!\bigl(\mathrm{SE}(\tilde{\mathbf{X}} - \mathbf{X})\bigr) \in \mathbb{R}^{C},
  \quad
  \mathbf{X}_{\mathrm{out}} = \mathbf{X} + \mathbf{g} \odot (\tilde{\mathbf{X}} - \mathbf{X}),
\end{equation}
where SE denotes a squeeze-and-excitation block. The gate selectively retains channels for which the predicted shift yields a meaningful correction, suppressing noisy offsets in spectrally homogeneous regions.

DADC replaces the standard depth-wise convolution inside the FFN of every Transformer block and is deployed at all scales of the U-Net denoiser, enabling coarse-to-fine dispersion correction. To tighten the coupling between the physics-based $\mathbf{x}$-step and the learned prior, PDAUN feeds the data-fidelity residual $\mathbf{r} = \mathbf{A}_{\text{disp}}^{T}(\mathbf{y}_{\text{disp}} - \mathbf{A}_{\text{disp}}\,\mathbf{x}_{k+1})$ as an additional spatial map to the denoiser, telling DADC where the current estimate still violates the forward model.

\subsection{Architecture Overview}

\cref{fig:architecture} illustrates the overall PDAUN pipeline. Each unfolding stage comprises three updates: a PCG-based data-fidelity step that projects the current estimate onto the feasible set of the ODIS forward model, a DADC-based prior step that acts as a learned denoiser under PAN guidance, and the dual-variable update that accumulates the constraint residual. Within the denoiser, each Transformer block first applies Attn-PH, a PAN--hyperspectral cross-modal attention module, and then a DADC-enhanced feed-forward network. Before unfolding begins, an initialization module performs two functions: (i)~a PAN feature pyramid extracts three-scale guidance maps $\mathbf{G}$ that are shared across all stages and scales of the U-Net denoiser; (ii)~a lightweight parameter-estimator predicts the stage-wise parameters $\{\rho_k,\lambda_k\}$ as well as scaling factors for the data terms, enabling content-adaptive balancing between spatial consistency and spectral consistency.

\section{Experiments}
\label{sec:experiments}

We evaluate ODIS and PDAUN through four sets of experiments. First, a cross-system comparison (\cref{sec:cross-system}) confirms the decisive advantage of ODIS's full light throughput for reconstruction under low illumination. Second, an algorithm-level comparison (\cref{sec:algo-compare}) demonstrates that PDAUN achieves state-of-the-art performance among existing reconstruction methods adapted to the ODIS forward model. Third, real-world experiments (\cref{sec:real-world}) on a physical prototype validate high-fidelity spectral recovery under both standard and low-light conditions. Finally, ablation studies (\cref{sec:ablation}) quantify the individual contributions of the FFT-Woodbury PCG solver and the DADC module.

\begin{figure}[tb]
  \centering
  \includegraphics[width=\linewidth]{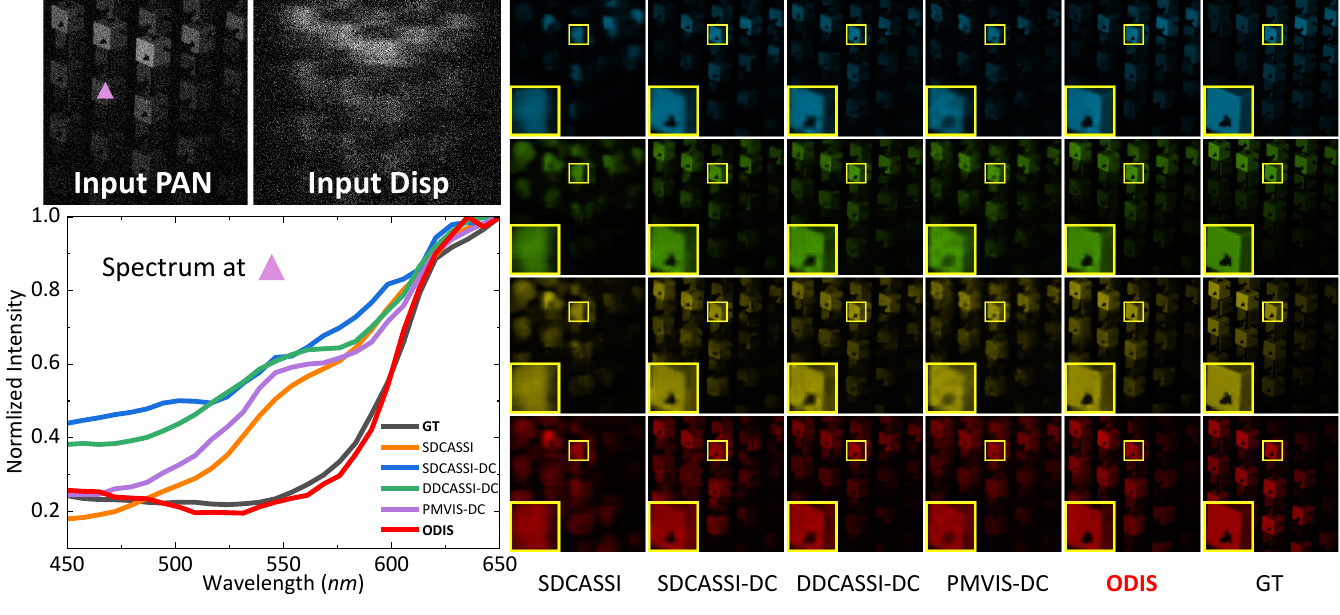}
  \caption{Cross-system simulation HSI reconstruction comparison under low illumination ($9$\,lux). The bottom-left illustrates normalized spectrum curves at the position marked by \emph{pink triangle} in the PAN image. The right shows reconstruction results at spectral bands 476.5nm, 536.5nm, 584.3nm, 648.1nm. Zoom in for a better view.}
  \label{fig:cross-system-vis}
  
  \captionof{table}{Cross-system comparison under multiple illumination levels.
    All systems use a unified U-Net backbone.
    Best results are in \textbf{bold}.}
  \label{tab:cross-system}
  \centering
  \resizebox{\textwidth}{!}{%
  \begin{tabular}{@{}l ccc @{\quad} ccc @{\quad} ccc @{\quad} ccc @{\quad} ccc@{}}
    \toprule
    \multirow{2}{*}{Lux}
    & \multicolumn{3}{c}{SDCASSI}
    & \multicolumn{3}{c}{SDCASSI-DC}
    & \multicolumn{3}{c}{DDCASSI-DC}
    & \multicolumn{3}{c}{PMVIS-DC}
    & \multicolumn{3}{c}{ODIS (Ours)} \\
    \cmidrule(lr){2-4} \cmidrule(lr){5-7} \cmidrule(lr){8-10} \cmidrule(lr){11-13} \cmidrule(lr){14-16}
      & PSNR{\color{green}$\uparrow$} & SSIM{\color{green}$\uparrow$} & SAM{\color{red}$\downarrow$}
      & PSNR & SSIM & SAM
      & PSNR & SSIM & SAM
      & PSNR & SSIM & SAM
      & PSNR & SSIM & SAM \\
    \midrule
    141
      & 28.82 & 0.8439 & \textbf{8.76}
      & 32.96 & 0.9406 & 10.37
      & 29.05 & 0.9039 & 13.93
      & 32.49 & 0.9316 & 11.68
      & \textbf{33.67} & \textbf{0.9504} & 9.62 \\
    35
      & 27.27 & 0.8018 & \textbf{10.33}
      & 31.27 & 0.9124 & 11.34
      & 27.38 & 0.8680 & 16.72
      & 29.14 & 0.8852 & 13.84
      & \textbf{32.79} & \textbf{0.9351} & 10.38 \\
    18
      & 26.36 & 0.7758 & 11.66
      & 30.16 & 0.8796 & 12.77
      & 26.81 & 0.8273 & 18.88
      & 26.21 & 0.7927 & 19.96
      & \textbf{32.29} & \textbf{0.9204} & \textbf{10.58} \\
    9
      & 25.21 & 0.7441 & 13.02
      & 28.44 & 0.8379 & 13.64
      & 26.27 & 0.8075 & 17.65
      & 25.63 & 0.7757 & 19.23
      & \textbf{30.67} & \textbf{0.8858} & \textbf{11.70} \\
    4
      & 23.82 & 0.7022 & 15.14
      & 26.62 & 0.7786 & 14.13
      & 25.22 & 0.7633 & 16.67
      & 25.14 & 0.7489 & 16.60
      & \textbf{29.10} & \textbf{0.8462} & \textbf{12.47} \\
    2
      & 22.86 & 0.6570 & 16.08
      & 24.62 & 0.7147 & 16.08
      & 23.78 & 0.7106 & 17.99
      & 20.19 & 0.4983 & 21.68
      & \textbf{27.03} & \textbf{0.7879} & \textbf{14.51} \\
    \bottomrule
  \end{tabular}}
\end{figure}

\subsection{Cross-System Comparison}
\label{sec:cross-system}

\noindent\textbf{Setup.}
To quantify the system-level throughput advantage of ODIS, we design a controlled cross-system simulation benchmark. Five representative CSI architectures are compared: (1)~standard single-path SD-CASSI (SDCASSI); (2)~SD-CASSI with a beam-splitter-coupled PAN camera (SDCASSI-DC); (3)~DD-CASSI with a beam-splitter-coupled PAN camera (DDCASSI-DC); (4)~PMVIS with a beam-splitter-coupled PAN camera (PMVIS-DC); and (5)~ODIS (proposed). For SDCASSI-DC, DDCASSI-DC, and PMVIS-DC, a 50\%/50\% beam splitter divides the incident light into a coded-measurement arm and a PAN arm; the beam-splitter loss is accounted for in each system's effective throughput. All dual-path systems share a unified U-Net backbone for reconstruction, ensuring that any observed performance gap is attributable to the physical information captured by each system rather than to algorithm-specific designs.

\noindent\textbf{Simulation datasets.}
Following the schedule of TSA-Net~\cite{meng2020end}, we trained all systems for 300 epochs on CAVE~\cite{park2007multispectral} dataset (28 channels). 10 scenes from KAIST~\cite{choi2017high} are selected for testing. The illumination-dependent noise model is applied to both training and evaluation to faithfully reflect each light condition.

\noindent\textbf{Simulation illumination model.}
Each measurement is corrupted by two noise sources: signal-dependent shot noise, whose severity is governed by the received photon count, and additive Gaussian noise caused by temporal dark noise of the sensor. The mapping from scene illuminance to noise parameters is jointly calibrated with a photometer and a FLIR camera (details in the supplementary material). Under this model, with a standard 500\,ms simulated exposure, illuminance levels of \{2, 4, 9, 18, 35, 141\}\,lux correspond to \{5, 6, 7, 8, 9, 11\}-bit shot noise with $\sigma_d = 9.29$\,e$^-$ Gaussian noise.

\noindent\textbf{Quantitative results.}
\cref{tab:cross-system} summarises PSNR, SSIM, and SAM across all systems and illumination levels. At high illumination (141\,lux), the raw measurement SNR of all systems is already sufficient for reconstruction: notably, SDCASSI achieves the best SAM (8.76) despite its 50\% throughput, because the coded mask substantially reduces the spectral ill-posedness of the inverse problem, yielding a slight advantage over ODIS (SAM 9.62). However, as illumination decreases, the full-throughput advantage of ODIS becomes decisive. The higher measurement SNR afforded by near-100\% throughput makes ODIS measurements far more informative under low photon counts. At 2\,lux, ODIS surpasses the second-best system (SDCASSI-DC), improving 2.41\,dB in PSNR and 0.073 in SSIM, while reducing SAM by 1.57; PMVIS-DC deteriorates to 20.19\,dB / 0.498 SSIM / 21.68 SAM, effectively failing. The PSNR gap between ODIS and SDCASSI-DC widens monotonically from 0.71\,dB at 141\,lux to 2.41\,dB at 2\,lux, confirming that throughput becomes the dominant factor governing reconstruction quality under photon-starved conditions.

\noindent\textbf{Visual analysis.}
\cref{fig:cross-system-vis} shows multi-band reconstruction results for a representative low-light scene (9\,lux). SDCASSI-DC, DDCASSI-DC, and PMVIS-DC exhibit severe artefacts, while SDCASSI barely resolves the scene structure. In contrast, ODIS maintains sharp spatial details and accurate spectral curves that closely match the ground truth, confirming the SNR and spectral fidelity advantages of full light throughput under low illumination.

\subsection{Algorithm Comparison}
\label{sec:algo-compare}

\noindent\textbf{Setup.}
To evaluate PDAUN against existing reconstruction algorithms under the ODIS system, we conduct an algorithm-level comparison on a standard simulation benchmark. The training and test sets follow the same configuration as in \cref{sec:cross-system}. Compared methods span four categories: (1)~Iterative method: SIFDB~\cite{zhao2019spectral}; (2)~E2E method: TSA-Net~\cite{meng2020end}, HDNet~\cite{hu2022hdnet}; (3)~Transformer method: MST~\cite{cai2022mask}, CST~\cite{cai2022coarse}, SPECAT~\cite{yao2024specat}; and (4)~Deep Unfolding method: DAUHST~\cite{cai2022degradation}, In2SET~\cite{wang2024in2set}. All competing methods are adapted to the ODIS forward model for a fair comparison: the coded mask is set to $\phi \equiv 1$ in the data-fidelity step and the PAN input is aligned to the ODIS observation geometry. Evaluation metrics include PSNR, SSIM, SAM, parameter count, and GFLOPs.

\begin{table}[tb]
  \caption{Quantitative comparison of reconstruction algorithms on the ODIS simulation benchmark.
    All deep unfolding methods are configured with 5 stages.
    Best results are in \textbf{bold} and second best are \underline{underlined}.}
  \label{tab:algo-compare}
  \centering
  \resizebox{\textwidth}{!}{%
  \begin{tabular}{@{}llccccc@{}}
    \toprule
    Method & Type & Params (M) & GFLOPs & PSNR{\color{green}$\uparrow$} & SSIM{\color{green}$\uparrow$} & SAM{\color{red}$\downarrow$} \\
    \midrule
    SIFDB~\cite{zhao2019spectral}         & Iterative      & --     & --     & 28.04          & 0.7714          & 19.45 \\
    TSA-Net~\cite{meng2020end}       & E2E            & 44.252 & 135.21 & 35.88          & 0.9661          & 7.89  \\
    HDNet~\cite{hu2022hdnet}         & E2E            & 2.368  & 154.79 & 34.62          & 0.9593          & 8.86  \\
    MST~\cite{cai2022mask}           & Transformer    & 3.665  & 16.74  & 36.41          & 0.9697          & 8.49  \\
    CST~\cite{cai2022coarse}           & Transformer    & 1.360  & 16.16  & 35.23          & 0.9631          & 9.04  \\
    SPECAT~\cite{yao2024specat}        & Transformer    & 0.286  & 11.45  & 35.17          & 0.9592          & 10.27 \\
    DAUHST~\cite{cai2022degradation}        & Deep Unfolding & 3.437  & 39.39  & 37.69          & 0.9753          & 7.55  \\
    In2SET~\cite{wang2024in2set}        & Deep Unfolding & 5.431  & 34.77  & \underline{37.98} & \underline{0.9776} & \underline{7.23} \\
    \midrule
    \textbf{PDAUN (Ours)} & Deep Unfolding & 6.790 & 64.22 & \textbf{38.52} & \textbf{0.9799} & \textbf{6.86} \\
    \bottomrule
  \end{tabular}}
\end{table}

\noindent\textbf{Quantitative results.}
\cref{tab:algo-compare} summarises the reconstruction performance. Our method PDAUN achieves the best results across all three metrics, reaching 38.52\,dB in PSNR, 0.9799 in SSIM, and 6.86 in SAM. Compared with In2SET, the current state of the art for DCCHI systems, PDAUN improves PSNR by 0.54\,dB, SSIM by 0.0023, and reduces SAM by 0.37, demonstrating the effectiveness of the FFT-Woodbury PCG solver and DADC module tailored to the ODIS forward model.

\begin{figure}[tb]
  \centering
  \includegraphics[width=\linewidth]{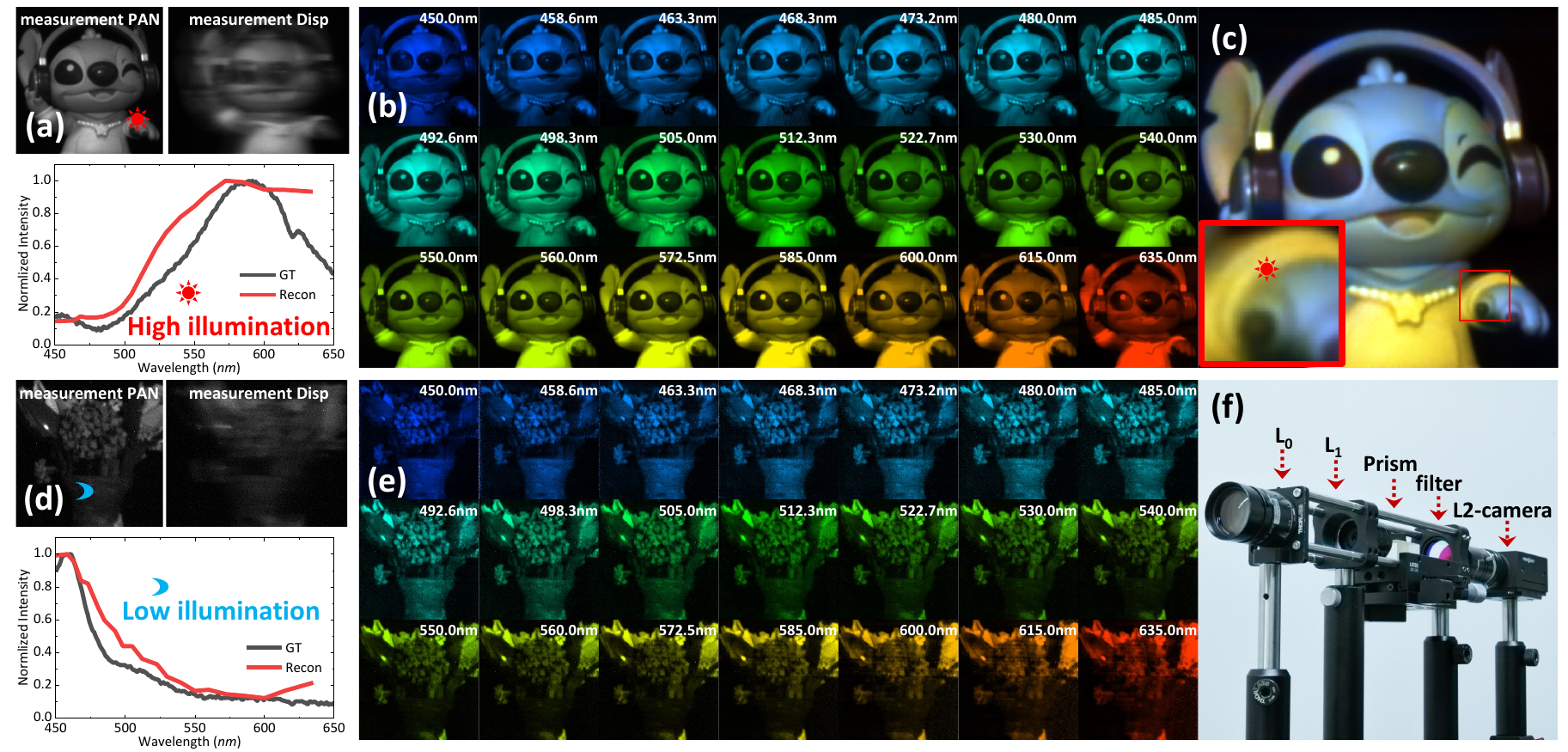}
  \caption{Real-world reconstruction on the ODIS prototype. (a)~Raw measurements under standard illumination ($\approx$774\,lux); spectral curves at the position marked by a \emph{red sun} are compared with spectrometer references. (b)~Reconstructed 21-channel HSI and (c)~its pseudo-RGB rendering. (d),\,(e)~Corresponding results under low-light conditions ($\approx$6.7\,lux). (f)~Photograph of the ODIS prototype.}
  \label{fig:real-world}
\end{figure}

\subsection{Real-World Experiment}
\label{sec:real-world}

\noindent\textbf{Prototype and acquisition.}
We build a physical ODIS prototype to validate spectral reconstruction under real optical conditions. As shown in \cref{fig:real-world}(f), the prototype consists of three lenses, an Amici dispersion prism, a linear translation stage, a monochrome camera (FLIR GS3-U3-41S4M-C), and a 450--650\,nm spectral filter. The prism oscillates between the image-conjugate plane and the defocused plane (travel $\approx$ 7\,mm, corresponding to an average dispersion step of $\approx$\,3\,pixels per 10\,nm) to acquire a PAN measurement and a dispersed measurement, respectively. We capture measurement pairs from multiple indoor scenes under both standard illumination ($\approx$\,774\,lux, simulating outdoor sunny-day conditions) and low-light conditions ($\approx$\,6.7\,lux, simulating twilight), with scene illuminance recorded by a calibrated light meter. For reconstruction, we train a two-stage variant (PDAUN-2stg) for 300 epochs on the CAVE and KAIST datasets, whose spectral channels are interpolated to 21 wavelengths spanning 450--650\,nm to match the prototype configuration. Illumination-dependent noise following the model in \cref{sec:cross-system} is injected during training.

\noindent\textbf{Standard-illumination reconstruction.}
The upper half of \cref{fig:real-world} presents reconstruction results under standard illumination ($\approx$\,774\,lux). PDAUN recovers a high-quality 21-band spectral data cube; the pseudo-RGB composite preserves natural colours and fine spatial details. To quantitatively assess spectral accuracy, we compare the reconstructed spectrum at the marked location in \cref{fig:real-world}(a) with a reference measurement obtained from a fiber-optic spectrometer (Avantes AvaSpec-Mini). The close agreement between the two curves confirms the spectral fidelity of the ODIS--PDAUN pipeline under real-world conditions.

\noindent\textbf{Low-light feasibility.}
The lower half of \cref{fig:real-world} shows reconstruction results under low-light conditions ($\approx$\,6.7\,lux). Despite the substantially reduced SNR, ODIS still yields recognisable spectral reconstructions that retain the principal spatial structures and spectral signatures of the scene. This preservation is attributable to the near-full throughput of ODIS, and corroborates the low-light advantage established in the simulation experiments.

\subsection{Ablation Study}
\label{sec:ablation}

\begin{table}[tb]
  \caption{Ablation studies.
    (a)~Break-down ablation on PDAUN's two core components:
    DADC (dispersion-aware deformable convolution) and
    FFT-WB PCG (FFT-Woodbury preconditioned CG).
    (b)~Effect of PCG iteration count.}
  \label{tab:ablation}
  \centering
  \footnotesize
  \begin{minipage}[t]{0.54\textwidth}
    \centering
    (a) Component ablation\\[3pt]
    \begin{tabular}{@{}lccccc@{}}
      \toprule
      Baseline & DADC & FFT-WB PCG & PSNR{\color{green}$\uparrow$} & SSIM{\color{green}$\uparrow$} & SAM{\color{red}$\downarrow$} \\
      \midrule
      \checkmark & --         & --         & 36.32          & 0.9684          & 8.77 \\
      \checkmark & --         & \checkmark & 36.47          & 0.9681          & 8.85 \\
      \checkmark & \checkmark & --         & 36.82          & 0.9702          & 8.07 \\
      \checkmark & \checkmark & \checkmark & 37.01 & 0.9736 & 8.09 \\
      \bottomrule
    \end{tabular}
  \end{minipage}
  \hfill
  \begin{minipage}[t]{0.34\textwidth}
    \centering
    (b) PCG iterations\\[3pt]
    \begin{tabular}{@{}lccc@{}}
      \toprule
      Iters & PSNR{\color{green}$\uparrow$} & SSIM{\color{green}$\uparrow$} & SAM{\color{red}$\downarrow$} \\
      \midrule
       5  & 37.01          & 0.9736          & 8.09 \\
       10 & 37.30          & 0.9743          & 7.76 \\
       15 & 37.31          & 0.9741          & 7.48 \\
       20 & 37.27 & 0.9745 & 7.01 \\
      \bottomrule
    \end{tabular}
  \end{minipage}
\end{table}

\noindent\textbf{Break-down ablation.}
We conduct a component-level ablation on PDAUN-2stg with 5 PCG (or CG) iterations to isolate the contribution of each proposed module (\cref{tab:ablation}a). The baseline, which replaces DADC with standard depth-wise convolution and FFT-Woodbury PCG with standard CG, yields 36.32\,dB in PSNR. Enabling DADC alone improves PSNR by 0.50\,dB, while enabling FFT-WB PCG alone contributes 0.15\,dB. The full model combining both modules achieves an improvement of 0.69\,dB in PSNR, 0.0052 in SSIM, and a decrease of 0.68 in SAM over the baseline, confirming the effectiveness of the proposed components.

\noindent\textbf{Effect of PCG iteration count.}
We further evaluate the impact of PCG iteration count on PDAUN-2stg (\cref{tab:ablation}b). Increasing the count from 5 to 10 yields a notable gain of 0.29\,dB in PSNR, indicating that additional iterations substantially improve spatial reconstruction quality. Beyond 10 iterations, PSNR plateaus while SAM continues to decrease from 7.76 to 7.01, demonstrating that further iterations primarily refine spectral fidelity.

\section{Conclusion}
\label{sec:conclusion}

We have presented ODIS, a compact single-camera system that achieves near-100\% light throughput by axially translating a dispersive prism along a single optical path without any coded mask or beam splitter. To reconstruct spectra from maskless ODIS observations, we develop PDAUN, an ADMM deep unfolding network integrating an FFT-Woodbury preconditioned CG solver with a Dispersion-Aware Deformable Convolution module for sub-pixel misalignment correction. Cross-system simulations confirm the decisive throughput advantage under low illumination; cross-algorithm comparisons show that PDAUN attains state-of-the-art performance; and high-fidelity reconstruction is validated on a physical prototype under standard ($\approx$\,774\,lux) and low-light ($\approx$\,6.7\,lux) conditions.

\noindent\textbf{Limitations and future work.}
ODIS requires two sequential exposures to acquire the dispersed measurement and the PAN image, partially sacrificing temporal resolution. Future work will introduce a high-speed motorised linear actuator for rapid prism oscillation and develop motion-artefact-aware reconstruction algorithms, jointly enabling video-rate spectral imaging.

\clearpage 

\bibliographystyle{splncs04}
\bibliography{main}

@String(CVPR  = {IEEE Conf. Comput. Vis. Pattern Recog.})

@String(AAAI  = {AAAI})

@String(ICIP  = {IEEE Int. Conf. Image Process.})

@String(PR    = {Pattern Recognition})

@String(CVPR  = {CVPR})

@String(ICIP  = {ICIP})

@String(PR    = {PR})

@article{bioucas2013hyperspectral,
  title={Hyperspectral remote sensing data analysis and future challenges},
  author={Bioucas-Dias, Jos{\'e} M and Plaza, Antonio and Camps-Valls, Gustavo and Scheunders, Paul and Nasrabadi, Nasser and Chanussot, Jocelyn},
  journal={IEEE Geoscience and remote sensing magazine},
  volume={1},
  number={2},
  pages={6--36},
  year={2013},
  publisher={IEEE}
}

@article{calin2014hyperspectral,
  title={Hyperspectral imaging in the medical field: Present and future},
  author={Calin, Mihaela Antonina and Parasca, Sorin Viorel and Savastru, Dan and Manea, Dragos},
  journal={Applied Spectroscopy Reviews},
  volume={49},
  number={6},
  pages={435--447},
  year={2014},
  publisher={Taylor \& Francis}
}

@article{lin2002reuven,
  title={The Reuven Ramaty high-energy solar spectroscopic imager (RHESSI)},
  author={Lin, Robert P and Dennis, Brian R and Hurford, Gordon J and Smith, DM and Zehnder, Alex and Harvey, PR and Curtis, David W and Pankow, Dave and Turin, Paul and Bester, M and others},
  journal={Solar Physics},
  volume={210},
  number={1},
  pages={3--32},
  year={2002},
  publisher={Springer}
}

@article{shimoni2019hyperspectral,
  title={Hyperspectral imaging for military and security applications: Combining myriad processing and sensing techniques},
  author={Shimoni, Michal and Haelterman, Rob and Perneel, Christiaan},
  journal={IEEE Geoscience and Remote Sensing Magazine},
  volume={7},
  number={2},
  pages={101--117},
  year={2019},
  publisher={IEEE}
}

@article{cao2016computational,
  title={Computational snapshot multispectral cameras: Toward dynamic capture of the spectral world},
  author={Cao, Xun and Yue, Tao and Lin, Xing and Lin, Stephen and Yuan, Xin and Dai, Qionghai and Carin, Lawrence and Brady, David J},
  journal={IEEE Signal Processing Magazine},
  volume={33},
  number={5},
  pages={95--108},
  year={2016},
  publisher={IEEE}
}

@article{lam2015computational,
  title={Computational photography with plenoptic camera and light field capture: tutorial},
  author={Lam, Edmund Y},
  journal={Journal of the Optical Society of America A},
  volume={32},
  number={11},
  pages={2021--2032},
  year={2015},
  publisher={Optical Society of America}
}

@article{yuan2021snapshot,
  title={Snapshot compressive imaging: Theory, algorithms, and applications},
  author={Yuan, Xin and Brady, David J and Katsaggelos, Aggelos K},
  journal={IEEE Signal Processing Magazine},
  volume={38},
  number={2},
  pages={65--88},
  year={2021},
  publisher={IEEE}
}

@article{arce2013compressive,
  title={Compressive coded aperture spectral imaging: An introduction},
  author={Arce, Gonzalo R and Brady, David J and Carin, Lawrence and Arguello, Henry and Kittle, David S},
  journal={IEEE Signal Processing Magazine},
  volume={31},
  number={1},
  pages={105--115},
  year={2013},
  publisher={IEEE}
}

@article{gehm2007single,
  title={Single-shot compressive spectral imaging with a dual-disperser architecture},
  author={Gehm, Michael E and John, Renu and Brady, David J and Willett, Rebecca M and Schulz, Timothy J},
  journal={Optics express},
  volume={15},
  number={21},
  pages={14013--14027},
  year={2007},
  publisher={Optical Society of America}
}

@article{wagadarikar2008single,
  title={Single disperser design for coded aperture snapshot spectral imaging},
  author={Wagadarikar, Ashwin and John, Renu and Willett, Rebecca and Brady, David},
  journal={Applied optics},
  volume={47},
  number={10},
  pages={B44--B51},
  year={2008},
  publisher={Optical Society of America}
}

@article{lapray2014multispectral,
  title={Multispectral filter arrays: Recent advances and practical implementation},
  author={Lapray, Pierre-Jean and Wang, Xingbo and Thomas, Jean-Baptiste and Gouton, Pierre},
  journal={Sensors},
  volume={14},
  number={11},
  pages={21626--21659},
  year={2014},
  publisher={MDPI}
}

@article{mihoubi2017multispectral,
  title={Multispectral demosaicing using pseudo-panchromatic image},
  author={Mihoubi, Sofiane and Losson, Olivier and Mathon, Benjamin and Macaire, Ludovic},
  journal={IEEE Transactions on Computational Imaging},
  volume={3},
  number={4},
  pages={982--995},
  year={2017},
  publisher={IEEE}
}

@article{wang2007concept,
  title={Concept of a high-resolution miniature spectrometer using an integrated filter array},
  author={Wang, Shao-Wei and Xia, Changsheng and Chen, Xiaoshuang and Lu, Wei and Li, Ming and Wang, Haiqian and Zheng, Weibo and Zhang, Tao},
  journal={Optics letters},
  volume={32},
  number={6},
  pages={632--634},
  year={2007},
  publisher={Optical Society of America}
}

@article{tittl2018imaging,
  title={Imaging-based molecular barcoding with pixelated dielectric metasurfaces},
  author={Tittl, Andreas and Leitis, Aleksandrs and Liu, Mingkai and Yesilkoy, Filiz and Choi, Duk-Yong and Neshev, Dragomir N and Kivshar, Yuri S and Altug, Hatice},
  journal={Science},
  volume={360},
  number={6393},
  pages={1105--1109},
  year={2018},
  publisher={American Association for the Advancement of Science}
}

@article{jeon2019compact,
  title={Compact snapshot hyperspectral imaging with diffracted rotation.},
  author={Jeon, Daniel S and Baek, Seung-Hwan and Yi, Shinyoung and Fu, Qiang and Dun, Xiong and Heidrich, Wolfgang and Kim, Min H},
  journal={ACM Trans. Graph.},
  volume={38},
  number={4},
  pages={117--1},
  year={2019}
}

@article{haraguchi2002spectral,
  title={Spectral imaging fluorescence microscopy},
  author={Haraguchi, Tokuko and Shimi, Takeshi and Koujin, Takako and Hashiguchi, Noriyo and Hiraoka, Yasushi},
  journal={Genes to Cells},
  volume={7},
  number={9},
  pages={881--887},
  year={2002},
  publisher={Wiley Online Library}
}

@article{fereidouni2012spectral,
  title={Spectral phasor analysis allows rapid and reliable unmixing of fluorescence microscopy spectral images},
  author={Fereidouni, Farzad and Bader, Arjen N and Gerritsen, Hans C},
  journal={Optics express},
  volume={20},
  number={12},
  pages={12729--12741},
  year={2012},
  publisher={Optical Society of America}
}

@article{miller2009dynamic,
  title={A dynamic lunar spectral irradiance data set for NPOESS/VIIRS day/night band nighttime environmental applications},
  author={Miller, Steven D and Turner, Robert E},
  journal={IEEE Transactions on Geoscience and Remote Sensing},
  volume={47},
  number={7},
  pages={2316--2329},
  year={2009},
  publisher={IEEE}
}

@inproceedings{kim2019deep,
  title={Deep learning based effective surveillance system for low-illumination environments},
  author={Kim, In Su and Jeong, Yunju and Kim, Seock Ho and Jang, Jae Seok and Jung, Soon Ki},
  booktitle={2019 Eleventh International Conference on Ubiquitous and Future Networks (ICUFN)},
  pages={141--143},
  year={2019},
  organization={IEEE}
}

@article{cao2011prism,
  title={A prism-mask system for multispectral video acquisition},
  author={Cao, Xun and Du, Hao and Tong, Xin and Dai, Qionghai and Lin, Stephen},
  journal={IEEE transactions on pattern analysis and machine intelligence},
  volume={33},
  number={12},
  pages={2423--2435},
  year={2011},
  publisher={IEEE}
}

@article{chen2023notch,
  title={A notch-mask and dual-prism system for snapshot spectral imaging},
  author={Chen, Linsen and Cai, Lijing and Huang, Erqi and Zhou, You and Yue, Tao and Cao, Xun},
  journal={Optics and Lasers in Engineering},
  volume={165},
  pages={107544},
  year={2023},
  publisher={Elsevier}
}

@article{wang2015dual,
  title={Dual-camera design for coded aperture snapshot spectral imaging},
  author={Wang, Lizhi and Xiong, Zhiwei and Gao, Dahua and Shi, Guangming and Wu, Feng},
  journal={Applied optics},
  volume={54},
  number={4},
  pages={848--858},
  year={2015},
  publisher={Optical Society of America}
}

@inproceedings{wang2015high,
  title={High-speed hyperspectral video acquisition with a dual-camera architecture},
  author={Wang, Lizhi and Xiong, Zhiwei and Gao, Dahua and Shi, Guangming and Zeng, Wenjun and Wu, Feng},
  booktitle={Proceedings of the IEEE conference on computer vision and pattern recognition},
  pages={4942--4950},
  year={2015}
}

@inproceedings{lv2023aperture,
  title={Aperture diffraction for compact snapshot spectral imaging},
  author={Lv, Tao and Ye, Hao and Yuan, Quan and Shi, Zhan and Wang, Yibo and Wang, Shuming and Cao, Xun},
  booktitle={Proceedings of the IEEE/CVF International Conference on Computer Vision},
  pages={10574--10584},
  year={2023}
}

@inproceedings{zhao2019spectral,
  title={Spectral reconstruction from dispersive blur: A novel light efficient spectral imager},
  author={Zhao, Yuanyuan and Hu, Xuemei and Guo, Hui and Ma, Zhan and Yue, Tao and Cao, Xun},
  booktitle={Proceedings of the IEEE/CVF Conference on Computer Vision and Pattern Recognition},
  pages={12202--12211},
  year={2019}
}

@inproceedings{cao2011high,
  title={High resolution multispectral video capture with a hybrid camera system},
  author={Cao, Xun and Tong, Xin and Dai, Qionghai and Lin, Stephen},
  booktitle={CVPR 2011},
  pages={297--304},
  year={2011},
  organization={IEEE}
}

@article{descour1995computed,
  title={Computed-tomography imaging spectrometer: experimental calibration and reconstruction results},
  author={Descour, Michael and Dereniak, Eustace},
  journal={Applied optics},
  volume={34},
  number={22},
  pages={4817--4826},
  year={1995},
  publisher={Optical Society of America}
}

@inproceedings{katkovnik2018optimization,
  title={Optimization of hybrid optics with multilevel phase mask for improved depth of focus broadband imaging},
  author={Katkovnik, Vladimir and Ponomarenko, Mykola and Egiazarian, Karen},
  booktitle={2018 7th European Workshop on Visual Information Processing (EUVIP)},
  pages={1--6},
  year={2018},
  organization={IEEE}
}

@inproceedings{yuan2016generalized,
  title={Generalized alternating projection based total variation minimization for compressive sensing},
  author={Yuan, Xin},
  booktitle={2016 IEEE International conference on image processing (ICIP)},
  pages={2539--2543},
  year={2016},
  organization={IEEE}
}

@article{liu2018rank,
  title={Rank minimization for snapshot compressive imaging},
  author={Liu, Yang and Yuan, Xin and Suo, Jinli and Brady, David J and Dai, Qionghai},
  journal={IEEE transactions on pattern analysis and machine intelligence},
  volume={41},
  number={12},
  pages={2990--3006},
  year={2018},
  publisher={IEEE}
}

@inproceedings{meng2020end,
  title={End-to-end low cost compressive spectral imaging with spatial-spectral self-attention},
  author={Meng, Ziyi and Ma, Jiawei and Yuan, Xin},
  booktitle={European conference on computer vision},
  pages={187--204},
  year={2020},
  organization={Springer}
}

@inproceedings{huang2021deep,
  title={Deep gaussian scale mixture prior for spectral compressive imaging},
  author={Huang, Tao and Dong, Weisheng and Yuan, Xin and Wu, Jinjian and Shi, Guangming},
  booktitle={Proceedings of the IEEE/CVF Conference on Computer Vision and Pattern Recognition},
  pages={16216--16225},
  year={2021}
}

@inproceedings{cai2022mask,
  title={Mask-guided spectral-wise transformer for efficient hyperspectral image reconstruction},
  author={Cai, Yuanhao and Lin, Jing and Hu, Xiaowan and Wang, Haoqian and Yuan, Xin and Zhang, Yulun and Timofte, Radu and Van Gool, Luc},
  booktitle={Proceedings of the IEEE/CVF conference on computer vision and pattern recognition},
  pages={17502--17511},
  year={2022}
}

@inproceedings{cai2022coarse,
  title={Coarse-to-fine sparse transformer for hyperspectral image reconstruction},
  author={Cai, Yuanhao and Lin, Jing and Hu, Xiaowan and Wang, Haoqian and Yuan, Xin and Zhang, Yulun and Timofte, Radu and Van Gool, Luc},
  booktitle={European conference on computer vision},
  pages={686--704},
  year={2022},
  organization={Springer}
}

@inproceedings{hu2022hdnet,
  title={Hdnet: High-resolution dual-domain learning for spectral compressive imaging},
  author={Hu, Xiaowan and Cai, Yuanhao and Lin, Jing and Wang, Haoqian and Yuan, Xin and Zhang, Yulun and Timofte, Radu and Van Gool, Luc},
  booktitle={Proceedings of the IEEE/CVF conference on computer vision and pattern recognition},
  pages={17542--17551},
  year={2022}
}

@article{cai2022degradation,
  title={Degradation-aware unfolding half-shuffle transformer for spectral compressive imaging},
  author={Cai, Yuanhao and Lin, Jing and Wang, Haoqian and Yuan, Xin and Ding, Henghui and Zhang, Yulun and Timofte, Radu and Gool, Luc V},
  journal={Advances in Neural Information Processing Systems},
  volume={35},
  pages={37749--37761},
  year={2022}
}

@inproceedings{wang2024in2set,
  title={In2set: Intra-inter similarity exploiting transformer for dual-camera compressive hyperspectral imaging},
  author={Wang, Xin and Wang, Lizhi and Ma, Xiangtian and Zhang, Maoqing and Zhu, Lin and Huang, Hua},
  booktitle={Proceedings of the IEEE/CVF Conference on Computer Vision and Pattern Recognition},
  pages={24881--24891},
  year={2024}
}

@inproceedings{gu2025hgsfusion,
  title={HGSFusion: Radar-camera fusion with hybrid generation and synchronization for 3D object detection},
  author={Gu, Zijian and Ma, Jianwei and Huang, Yan and Wei, Honghao and Chen, Zhanye and Zhang, Hui and Hong, Wei},
  booktitle={Proceedings of the AAAI Conference on Artificial Intelligence},
  volume={39},
  number={3},
  pages={3185--3193},
  year={2025}
}

@article{loncan2015hyperspectral,
  title={Hyperspectral pansharpening: A review},
  author={Loncan, Laetitia and De Almeida, Luis B and Bioucas-Dias, Jos{\'e} M and Briottet, Xavier and Chanussot, Jocelyn and Dobigeon, Nicolas and Fabre, Sophie and Liao, Wenzhi and Licciardi, Giorgio A and Simoes, Miguel and others},
  journal={IEEE Geoscience and remote sensing magazine},
  volume={3},
  number={3},
  pages={27--46},
  year={2015},
  publisher={IEEE}
}

@inproceedings{yao2024specat,
  title={Specat: Spatial-spectral cumulative-attention transformer for high-resolution hyperspectral image reconstruction},
  author={Yao, Zhiyang and Liu, Shuyang and Yuan, Xiaoyun and Fang, Lu},
  booktitle={Proceedings of the IEEE/CVF Conference on Computer Vision and Pattern Recognition},
  pages={25368--25377},
  year={2024}
}

@inproceedings{park2007multispectral,
  title={Multispectral imaging using multiplexed illumination},
  author={Park, Jong-Il and Lee, Moon-Hyun and Grossberg, Michael D and Nayar, Shree K},
  booktitle={2007 IEEE 11th International Conference on Computer Vision},
  pages={1--8},
  year={2007},
  organization={IEEE}
}

@techreport{choi2017high,
  title={High-quality hyperspectral reconstruction using a spectral prior},
  author={Choi, Inchang and Kim, Min Hyung and Gutierrez, Daniel and Jeon, Dong Seon and Nam, Gunhee},
  year={2017}
}
\end{document}